\documentclass[runningheads]{llncs}

\usepackage{eccv}

\usepackage{eccvabbrv}

\usepackage{graphicx}
\usepackage{booktabs}

\usepackage[accsupp]{axessibility}  

\usepackage{hyperref}

\usepackage{orcidlink}
\usepackage{multirow}
\usepackage{adjustbox}
\usepackage{
}
\usepackage{pifont}
\usepackage{makecell}
\usepackage[symbol]{footmisc}

\begin{document}

\title{HaineiFRDM: Structure-Preserving Diffusion for Film Restoration under Fast Motion and Diverse~Defects}


\author{Rongji Xun\inst{1}$^*$ \and
	Junjie Yuan\inst{2} \and
	Zhongjie Wang\inst{1}$^\dagger$}
\footnotetext[1]{This work was conducted during the author’s graduate studies.}
\footnotetext[2]{Corresponding author.}

\authorrunning{R.~Xun, J.~Yuan and Z.~Wang}

\institute{Tongji University, Shanghai, China \and
Shanghai Film Restoration Laboratory, Shanghai, China \\
\email{rongji.xun@gmail.com},
\email{sftp\_yuanjunjie@outlook.com},
\email{wang\_zhongjie@tongji.edu.cn},
}

\maketitle

\begin{abstract}
Existing film-restoration methods frequently fail under fast motion, producing limb disappearance and structural distortion due to inaccurate motion modeling. 
Moreover, high-resolution restoration under spatially-persistent and mixed defects remains insufficiently studied.
We propose HaineiFRDM, a Film Restoration Diffusion Model that leverages the content modeling capability of diffusion models for content-aware restoration, removing defects while preserving scene structure.
To enable scalable high-resolution restoration, we adopt a patch-wise strategy with position-aware global fusion modules to maintain cross-patch coherence. We further introduce a frequency-based module to enhance texture consistency and a patch-consistent inference framework to alleviate blocking artifacts introduced by patch-based processing.
We also construct a film restoration dataset comprising categorized defect templates, professionally restored films, and realistic synthetic degradations.
Extensive experiments demonstrate our superior restoration quality with strong structural consistency. Our design also reduces memory requirements, enabling high-resolution restoration on a single 24GB-VRAM GPU.
Code and the dataset will be released at
\url{https://anonymous.4open.science/r/HaineiFRDM}.
	
\end{abstract}    
\section{Introduction}
\label{sec:intro}

For film digital restoration, professionals typically manually restore every frame with the help of restoration software. This restoration process is labor-intensive and time-consuming. For instance, a professional usually only restore $200-2K$ frames per day, while a 2-hour film contains roughly $200K$ frames, requiring months of work for a restoration team.
The workload becomes heavier for films with widespread tricky defects, where professionals must repeatedly try different tools to achieve theater-quality results. Therefore, there exists urgent needs of developing restoration models to alleviate the restoration burden.

\begin{figure}[t]
	\centering
	\includegraphics[width=0.9\linewidth]{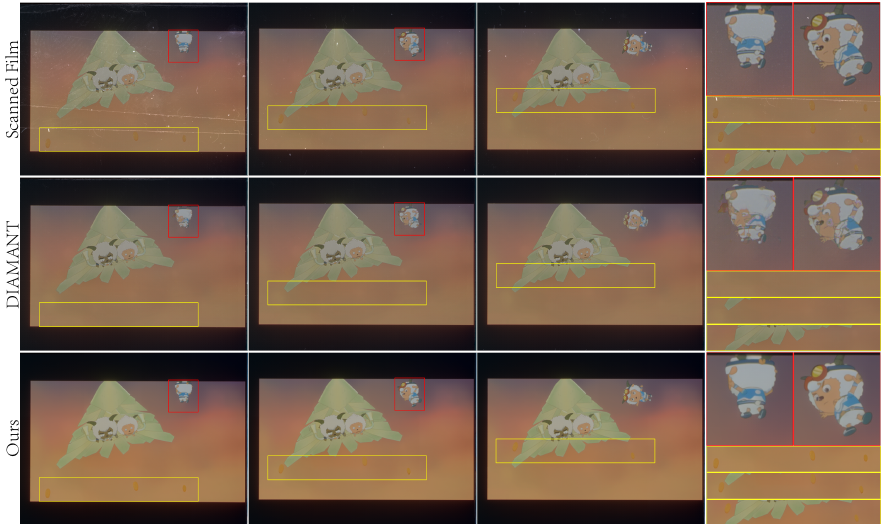}
	\caption{
		\textbf{Restoration results of real films with fast motion}: we display 3 consecutive 2K-resolution(2048*1536) frames. Compared with the DIAMANT\cite{DIAMANT}, our model maintains structure of fast-motion contents like `flying beans' and body.
	}
	\label{fig:real-xiyangyang}
	\vspace{-1.5em}
\end{figure}
The objective of professional restoration differs from the optimization target of many academic methods. Professionals favor removing most defects without introducing new artifacts and keeping clean regions unchanged. In contrast, many open-source approaches\cite{iizuka2019deepremaster,wan2022bringing,lin2024restoring,jiang2024deepenhancer,mao2025making} trained on synthetic degradation data, tend to over-restore frames with aggressive enhancement(\eg deblurring and contrast-boosting), which may alter the directors' aesthetic idea and limit their practical usability.
Beyond this objective mismatch, open-source and commercial methods still struggle on some challenging scenarios(noisy film with fast motion and constant scratch), often causing under-restoration and structural failures(object missing and distortion, as shown in \cref{fig:real-xiyangyang}) due to their reliance on inaccurate optical flows and imprecise alignment.
Besides, open-source methods generally lag behind commercial software on real films, largely due to the domain gap between synthetic degradations and real film defects, as they are typically trained with limited degradation diversity and without movie data.
Furthermore, high-resolution($\geq2K$-resolution) color film restoration is rarely considered in prior open-source works: direct RGB patch-wise processing often introduces blocking artifacts and cross-region inconsistency, as shown in \cref{fig:motivations}(b), making practical deployment difficult. These limitations motivate a restoration model that preserves structure under fast motion and remaining scalable to high-resolution films.

To address these gaps, we propose \textbf{HaineiFRDM}, a Film Restoration Diffusion Model for structure-preserving restoration under fast motion and complex defects. Unlike flow-guided methods that are easily confused by large motion, HaineiFRDM leverages diffusion-based content modeling to perform content-aware restoration: it firstly captures the underlying scene content, identifies and removes the film defects with the help of learned priors, and finally restore the content texture while preserving true structures. Thereby, this design maintains spatial and temporal consistency and reduces structural distortion like limb disappearance.
Guided by observations of degraded real films, we construct a film restoration dataset, covering movie clips with diverse film-shooting techniques and common film defect patterns, to mitigate the lack of public film restoration data. 
To make high-resolution restoration practical on commodity GPUs(typically $24-32$GB GPU in restoration labs), we further develop a patch-wise training strategy together with a Patch-Consistent Inference Framework that incorporates global residual guidance\cite{du2024demofusion,lin2025accdiffusion} to preserve cross-region coherence while reducing memory requirements and enabling $2K$-resolution film restoration on a single 24GB GPU. In addition, we introduce a frequency-based module to improve texture consistency of restored films.

\textbf{Our main contributions can be summarized as follows:}
1) We propose HaineiFRDM, a diffusion-based framework for structure-preserving film restoration under fast motion and complex defects;
2) We construct a film restoration dataset comprising professionally restored film pairs and realistic synthetic data;
3) We develop a scalable patch-wise training and Patch-Consistent Inference Framework that enables $2K$-resolution restoration on a single 24GB GPU, together with a frequency-based module to improve texture consistency.

\section{Related Works}\label{sec:relatedwork}
Film degradations can be broadly grouped into structured defects (\eg scratches, dust, abrasion) and unstructured degradations (\eg blur, noise, color fading).
Traditional methods \cite{saito1999image,kokaram2004missing, kim2006efficient} mainly target structured defects with a detect-then-restore pipeline, which helps preserve flawless regions while removing defects.
However, their reliance on hand-crafted features, simplifying assumptions, and heuristic thresholds limits their robustness on real films, especially for films with fine textures, large defects and fast motion, often leading to undetected defects and inconsistent textures.
Recent learning-based methods aim to handle both structured and unstructured degradations, and can be categorized by temporal modeling into implicit motion-modeling methods\cite{iizuka2019deepremaster,agnolucci2024reference} and flow-guided methods\cite{wan2022bringing,lin2024restoring,jiang2024deepenhancer,mao2025making}.

\noindent{\textbf{Implicit motion-modeling methods}} learn spatiotemporal features without explicit motion cues.
DeepRemaster\cite{iizuka2019deepremaster} adopts 3D convolutions, but its limited receptive field can leave defect residuals and cause temporal inconsistency. 
TAPE\cite{agnolucci2024reference} employs attention-based feature propagation, improving restoration quality, yet it still struggles to disambiguate defects from complex textures;

\noindent{\textbf{Flow-guided methods}} often built upon the BasicVSR-style propagation framework~\cite{chan2021basicvsr}, use optical flow to align and propagate information across frames for recognizing defects and improving temporal consistency~\cite{wan2022bringing,lin2024restoring,jiang2024deepenhancer,mao2025making}.
Some works\cite{jiang2024deepenhancer,mao2025making} further adopt efficient attention mechanisms\cite{yang2021focal} or sequence models\cite{gu2024mamba} to enhance spatio-temporal modeling.
However, optical flow remains \mbox{unreliable} under large motion or complex defects, which can mislead restoration to mistakenly remove true content as defects, which is unacceptable in professional workflows.
Moreover, most methods rely on limited and less realistic synthetic degradations, resulting in a significant domain gap to real films.

Overall, existing methods have limited content-awareness and suffer from structural distortion and incomplete defect removal, and are rarely evaluated for high-resolution real films with diverse motion and complex defects. These limitations motivate content-aware and scalable film restoration frameworks.

\noindent{\textbf{High-resolution Generation Diffusion}}~~
To lower GPU memory of high-resolution generation, high-resolution image generation methods explore patch-wise generation and global-to-local guidance.
MultiDiffusion\cite{bar2023multidiffusion} adopts a sliding-window scheme to generate and merge patches, but may suffer from repetition artifacts.
Besides, several training-free approaches first synthesize a low-resolution image and then upscale during denoising.
ScaleCrafter \cite{he2023scalecrafter} and FouriScale\cite{huang2024fouriscale} leverage dilated convolutions for feature upsampling, yet can struggle with texture consistency across patches.
DemoFusion\cite{du2024demofusion} introduces global residual guidance by generating a low-resolution prediction and using its upsampled features to guide patch-wise denoising, inspiring subsequent variants\cite{lin2025accdiffusion, tragakis2024one, jeong2025latent}. Other works\cite{yang2025inf,skorokhodov2024hierarchical} redesign network architectures and inject explicit patch-position cues to improve high-resolution coherence.

\section{Method}\label{sec:Method}
\begin{figure}[t]
	\centering
	\includegraphics[width=1.0\linewidth]{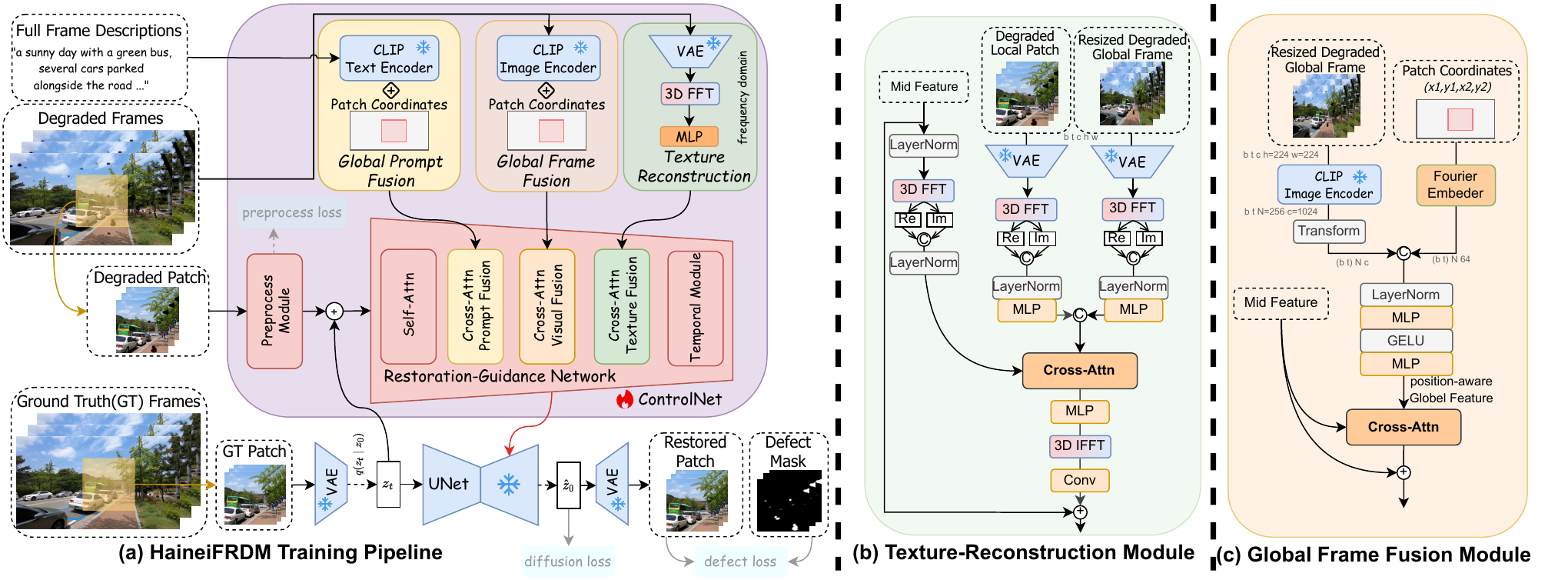}
	\caption{
		\textbf{Training pipeline of our proposed HaineiFRDM.} (a) patch-based training pipeline of HaineiFRDM;
		(b)the detailed structure of proposed texture reconstruction module;
		(c)the detailed structure of proposed global frame fusion module.
	}
	\label{fig:TrainFramework}
	\vspace{-0.5em}
\end{figure}
To perfectly remove defects across diverse restoration scenarios, our HaineiFRDM leverages the content modeling and generation ability of diffusion models for content-aware restoration:  firstly capture scene content, then removes defects, and thus largely reducing structural distortions.

Given the high-resolution nature of practical films, we firstly design a \textbf{patch-based training framework}(\cref{sec-Training-Framework}) to make model training and high-res films restoration possible on 24GB GPUs. In \cref{sec-Training-Framework}, we introduce \textbf{global frame fusion module} to compensate for the limited context of local patches, design a \textbf{global prompt fusion module} to better exploit prompt-conditioned diffusion modeling and also devise a \textbf{texture-reconstruction module} to help model better maintain texture through frequency domain.
Secondly, we design a \textbf{patch-consistent inference framework} in \cref{sec-infer-framework}, to further reduce memory usage and mitigate the unacceptable blocking artifacts in 2K-resolution restoration.

\subsection{Patch-based Training Framework}\label{sec-Training-Framework}
To lower the memory costs of high-resolution restoration, we design a patch-based training framework as shown in \cref{fig:TrainFramework}.
Following shifted-window mechanism\cite{liang2022recurrent}, we split training full video frames into overlapped video-patches, randomly sample a patch position to enrich data diversity and then input the model with the corresponding degraded patch, GT(Ground Truth) patch, full-frame description and defect masks.
Furthermore, we encode the degraded patch with our Preprocess Module to obtain a preprocessed feature, and use VAE model to encode the GT patch into latent space to get GT feature, thus further lowering memory cost.
Then we randomly sample a timestep $t$, add noise to GT patch's VAE feature to get noisy feature $z_t$, fuse the noisy GT feature and the preprocessed feature and input the feature into Restoration-Guidance Network.
In Restoration-Guidance Network, the features would sequentially processed by self-attrntion, global-prompt-fusion, global-frame-fusion, texture-reconstruction and temporal modules. Especially,the global-prompt and global-frame fusion modules enhance global awareness when restoring a local patch, while the texture-reconstruction module improves texture consistency. These modules help generate learned restoration guidance, which is injected into UNet to produce restored frame features $\hat{z_0}$. Finally, we decode the $\hat{z_0}$ into RGB frames and introduce a defect loss to highlight defects region in restored frames.

\begin{figure}[t]
	\centering
	\includegraphics[width=1.0\linewidth]{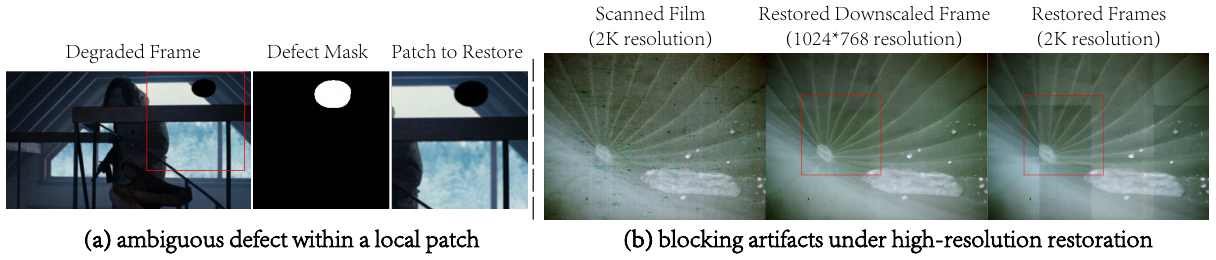}
	\caption{
		\textbf{Motivations of our proposed modules.}}
	\label{fig:motivations}
	\vspace{-1.5em}
\end{figure}

\noindent{\textbf{Global Fusion Module:}}
Though restoring each patch separately lowers the memory usage, a local patch may have inadequate information for the model to distinguish defects with flawless content, which results in content-distortion and false-restoration problem when restoring high-resolution and fast-movement frames.
For example, if merely observing the cropped patch frames showed in \cref{fig:motivations}(a), black defects have high similarities with hand rail in the frame and patch information is not enough for model to recognize defects correctly.

To solve the problem, we design global-frame-fusion(\cref{fig:TrainFramework}(c)) and a global-prompt-fusion modules to increase global awareness of the model. 
For global-frame-fusion module, we fuse bounding box coordinates of the patch and global-frame to get a location-related visual feature and then inject the feature into the model via cross-attention mechanism, following object-location-control methods\cite{ma2024subject, xiao2025fastcomposer}.
Specifically, we firstly resize the degraded global frames and feed each frame into CLIP\cite{radford2021learning} image encoder to get visual patch tokens that bearing more positional information than CLS token.
Secondly, we transform the patch coordinates into location features via Fourier Embeder and fuse the location feature with the visual patch tokens to get a position-aware global frame feature. Finally, we use cross-attention mechanism to fuse the global-frame feature with mid features outputted from previous module and then inject the fused feature back into the Restoration-Guidance Network.

For global-prompt-fusion module, we adapt the full-frame description to fully exploit prompt-conditioned modeling of diffusion models. One idea is to generate textual descriptions of every patches using LLMs\cite{hong2024cogvlm2}, which is time-costing for high-resolution and long-duration films. Therefore, we propose to generate full frame's video descriptions and exploit the textual information via a similar procedure used in the global-frame-fusion module.

\noindent{\textbf{Texture-reconstruction Module:}}
Consistent film texture is essential for practical film restoration.
As shown in \cref{fig:ablation-freqModule}, though the global-frame-fusion module helps improving content-consistency of restored frames, high-frequency texture of restored frames still have obvious texture-mismatch problem. 

To improve texture consistency, we following SFHformer\cite{jiang2024fast} to restore texture in frequency domain and propose a frequency-based texture-reconstruction module, depicted in \cref{fig:TrainFramework}(b).
Specifically, we firstly extract a VAE feature of the degraded local patch and transform the feature into freuqncy domain using 3D-FFT. Meanwhile, we extract frequency feature of resized degraded global frames and use it as reference information to maintain low-frequency information(\eg, color and lightness) of original frames. 
After, we concatenate the real and image part of 3D-FFT result and use MLP to further optimize the frequency features.
Lastly, we transform the model mid feature into frequency feature, fuse the frequency feature with cross-attention, then use 3D-iFFT to transform the frequency feature back into feature space and inject to the model.

\subsection{Patch-Consistent Inference Framework}\label{sec-infer-framework}
\begin{figure}[t]
	\centering
	\includegraphics[width=1.0\linewidth]{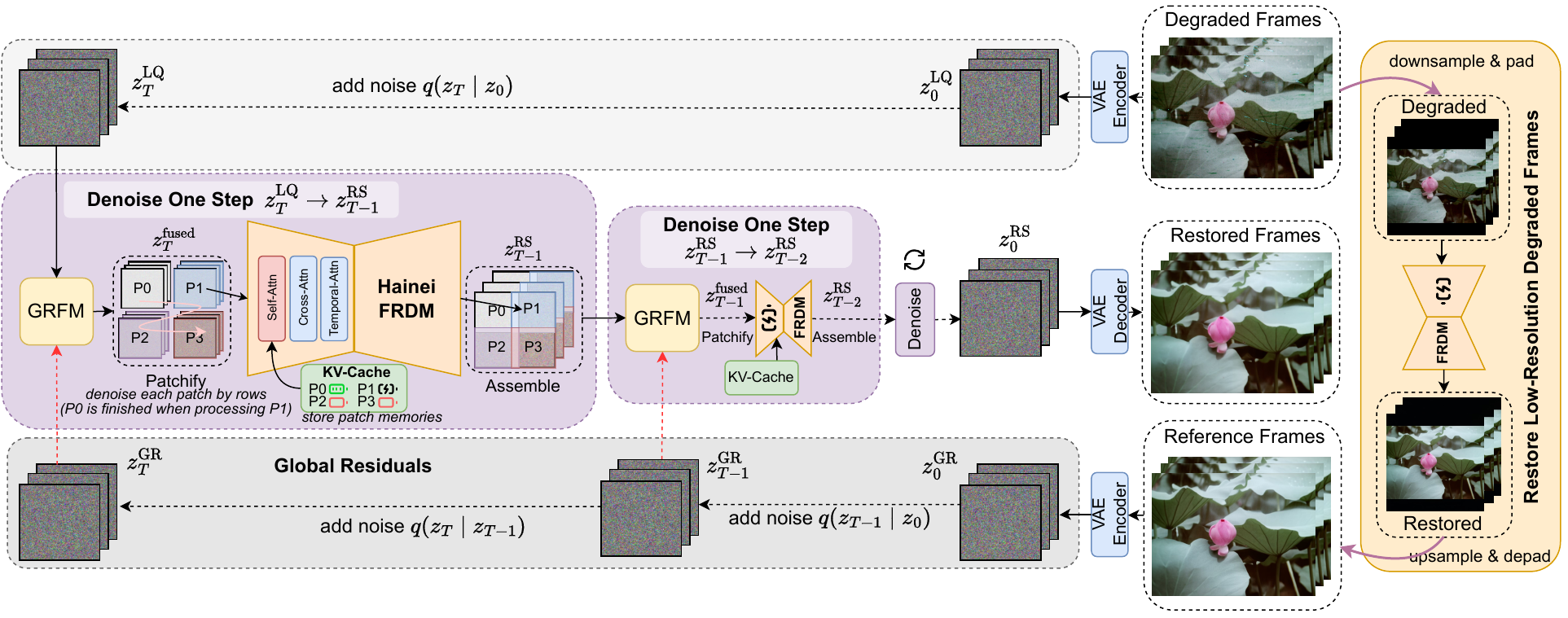}
	\caption{
		\textbf{Patch-Consistent Inference Framework}}
	\label{fig:patch-inference-framework}
	\vspace{-1em}
\end{figure}
As shown in \cref{fig:motivations}(b), applying the above modules and restoring RGB patches independently can yield relatively consistent results at low resolutions.
However, at theater-level resolutions ($\geq$2K), a same-sized patch contains much less global context: 1)global-level invisible film grains becomes indistinguishable from dense defects;2) local brightness can be dominated by large-area defects, leading the model produce color shifts. Moreover, independent patch restoration lacks cross-patch correlation, which further causes texture inconsistency across patches.

To solve blocking artifacts, we following high-resolution image generation methods\cite{bar2023multidiffusion, du2024demofusion, yang2025inf} and design a patch-consistent inference framework, depicted in \cref{fig:patch-inference-framework}, a two-stage framework employing global residuals\cite{du2024demofusion} to maintain full-frame consistency.
Specifically, at first stage, we firstly downsample and pad the high-resolution frames to a patch-sized frames, restore it with our HaineiFRDM and upsample the frames back to original resolution.
This pre-restored frame removes most defects while preserving global content and color consistency, making it a suitable reference for guiding local patches. We therefore extract features from the reference frame and add noise to obtain global residual latents $z^{GR}_{t}$, which providing reference information for each denoising timestep.
After that, we extract features for original degraded frames and add noise to get a noisy low-quality feature $z^{LQ}_T$ for second stage(denoising).

In each denoising timstep $t$, we firstly fuse the previous-timestep restored feature $z_t^{RS}$ with the global residual feature $z^{GR}_{t}$ using GRFM(Global Residual Fusion Module) to get a fused feature $z_t^{fused}$. The GRFM is implemented as a re-weighting model as \cref{eq:infer_cos_fuse}.
\begin{equation}
	\label{eq:infer_cos_fuse}
	z_t^{fused} = (1 - c_t) \cdot z_t^{RS} + c_t \cdot z_t^{GS}
\end{equation}
\begin{equation}
	\label{eq:infer_cos_weight}
	c_t = 0.25 \cdot \left[ 1 + cos(\pi \cdot \frac{T-t}{T}) \right],
\end{equation}
where we use $z^{LQ}_T$ as $z_T^{RS}$ at timestep $T$. And we use a cosine-decreasing weight $c_t$ for the global residuals, as in \cref{eq:infer_cos_weight}. Furthermore, we patchify the fused feature $z^{fused}_t$, denoise each patch with our HaineiFRDM model respectively and assemble the patches to get a restored feature $z_{t-1}^{RS}$.
Lastly, we repeat the above denoising process until getting the final restored frames' features $z_0^{RS}$ and use VAE Decoder to transform the feature back to RGB restored frames.

Besides the global residual, we also employ KV-cache\cite{yang2025inf} mechanism in self-attention module to pass previous patches restoration memory to HaineiFRDM, helping model better maintain texture consistency among neighboring patches.

\subsection{Loss Functions}
We firstly use \textbf{Diffusion Loss} $\mathcal{L}_{noise}$ to guide the training of diffusion model.
\begin{equation}
	\label{eq:loss1}
	\mathcal{L}_{noise}=\mathbb{E}_{z_0,t, \epsilon \sim \mathcal{N}(0,1)}\left[
	\left\|\epsilon-\epsilon_{\theta}\left(z_t,t\right)\right\|_2^2
	\right],
\end{equation}
where $t$ is a radom-sampled timestep, $\epsilon$ is a random noise, $\epsilon_{\theta}$ denotes a noise predicted by diffusion model(parameterized by $\theta$). And $z_0$ is flawless frames features, $z_t$ denotes previous-timestep feature.

\noindent{\textbf{Preprocess Loss}}~~
To extract defect-irrelevant frame features, we follow \cite{yang2024pixel} to extract frame features in multiple scales and then transform the feature back to RGB space to compute loss, 
\begin{equation}
	\label{eq:loss2}
	\mathcal{L}_{preprocess} = \sum_{j=1}^{S}\left[
		\frac{1}{N}\sum_{i}^{N}
		\left\|
		I_{preprocess}^{i,j} - I_{GT}^{i} 
		\right\|_1
	\right],
\end{equation}
where $N$ is frames number, $S$ is scale number, $I_{preprocess}$ denotes frame predicted by the Preprocess Module and $I_{GT}$ denote flawless frame.

\noindent{\textbf{Defect Loss}}~~
The losses above mainly supervise full-frame reconstruction quality. Since film defects often occupy only a small fraction of pixels, their gradients can be overwhelmed by full-frame losses, making the model less effective at removing small defects. To emphasize defect pixels, we introduce the defect loss, decoding the restored feature $\hat{z_0}$ into RGB frame $I_{pred}$ and then only computing pixel-wise loss over defect regions using the defect masks $m_{defect}$.
\begin{equation}
	\label{eq:loss3}
	\mathcal{L}_{defect} = 	\frac{1}{N}\sum_{i=1}^{N}
	\left\|
	(I_{pred}^{i} - I_{GT}^{i}) \cdot m_{defect}^{i}
	\right\|_1
	,
\end{equation}
The above losses form our final optimization goal:
\begin{equation}
	\label{eq:loss-all}
	\mathcal{L}_{total} =
	 	\mathcal{L}_{noise} + 
	 	\alpha_{p} \mathcal{L}_{preprocess} + 
	 	\alpha_{d} \mathcal{L}_{defect}
	.
\end{equation}
where we empirically set $\alpha_{p}=1$ and $\alpha_{d}=81$.

\section{Film Restoration Dataset}
\label{sec:film_dataset}
\begin{figure}
	\centering
	\includegraphics[width=1.0\linewidth]{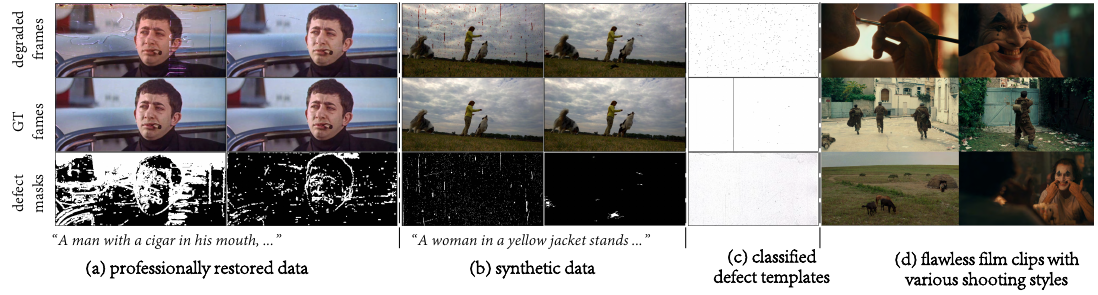}
	\caption{The composition of our Film Restoration Dataset.}
	\label{fig:data}
	\vspace{-1.5em}
\end{figure}

Due to lacking open-source high-quality restoration data, we constructed a dataset that comprising synthetic data and professionally restored real films.
And each data in our dataset is composed of degraded frames, restored frames, defects masks and video description, as showed in \cref{fig:data}.

\noindent{\textbf{Real Film Data:}} 
The degraded frames are obtained from scanned films, and the corresponding flawless frames are restored by professionals.
To construct restored films pairs, we collected before/after restoration videos released by restoration laboratories and commercial vendors, including MTI\cite{MTI}, DIAMANT\cite{DIAMANT},VIVA-pro\cite{VIVA-pro} and FilmFinity\cite{Filmfinity}. 
However, some restored videos may have pixel-mismatch problems and defects remains. We therefore extracted degraded and restored frames from the videos, and manually evaluate samples.
For the qualified pairs, we compute binary defect masks from the paired frames, manually label video captions or generate captions using CogVLM2\cite{hong2408cogvlm2}.
Combining the web-collected pairs and laboratory-restored films, we obtain approximately 5 minutes(24fps) of paired real-film data for training, and collect 10-minute-duration 2K-resolution real films for testing.

\noindent{\textbf{Synthetic Data:}}
To increase data diversity in both volume and degradation types, we synthesized degradation data under professional guidance. Specifically, we collected flawless film clips with different shooting styles (\cref{fig:data}d) and curated defect templates with different degradation types (\cref{fig:data}c), including sparse and intensive dust, cigarette-burn holes, flickering and constant scratches, \etc. Furthermore, we randomly apply quality degradation(rescale,jpeg compression) and film texture simulation(film grain) to the flawless frames, vary the color of defects in templates and fuse the flawless frames and the defects to obtain the final degraded frames.

%

\section{Experiments}
\label{sec:experiments}
\subsection{Experimental Setup}
We train our HaineiFRDM for $500K$ iterations on four NVIDIA 4090-24GB GPU and evaluate on one 24GB-VRAM GPU. We use the pretrained Stable Diffusion V1.5\cite{rombach2022high} and the motion module of AnimateDiff\cite{guo2023animatediff} for initialization, and freeze the CLIP\cite{radford2021learning}, VAE and UNet model during training.

\noindent{\textbf{Baseline:}} We compare our method with both the commercial software DIAMANT\cite{DIAMANT} and open-source methods(DeepRemaster\cite{iizuka2019deepremaster}, RTN\cite{wan2022bringing}, RRTN\cite{lin2024restoring}, DeepEnhancer\cite{jiang2024deepenhancer}).

\noindent{\textbf{Test Data:}} We evaluate models on both real films and synthetic data covering three practical degradation settings: mixed degradation, scratch degradation and noisy fast-motion films. For mixed degradations, we pick 6 DAVIS\cite{pont20172017} video with diverse motion patterns and randomly apply classified defect templates onto each frame to simulate severely degraded films.
For scratch degradation, we pick another 6 DAVIS\cite{pont20172017} video with diverse motion and apply collected scratch templates to simulate the challenging fixed-region scratch degradation.
For noisy fast-motion films, we synthesize realistic degradations on 2K-resolution cartoon sequences under professional guidance to evaluate models’ ability to recognize ambiguous defects and maintain structural consistency. For real films, we collect 2K-resolution scanned films, covering various motion patterns.

\noindent{\textbf{Metrics:}}
We report per-frame quantitative metrics to evaluate restoration quality.
We use PSNR and SSIM\cite{wang2004image} to evaluate each frame's overall restoration accuracy.
However, common film defects only take up small fractions of pixels and so the full-frame PSNR metric is often insensitive to reflect defect pixels' restoration performance. Therefore, we introduce a metric MaskedPSNR$^*$, which calculates the PSNR metric of defects pixels by using collected binary mask:
\footnotetext[1]{We abbreviate MaskedPSNR as MPSNR in qualitative figures to save sapce.}
\begin{equation}
\label{eq:MaskedPSNR}
	\left\{\begin{matrix}  
	 MSE_{mask} = \frac{1}{\sum{m_{defect}}} \cdot
	 	{
	 		\left[ \left(  I_{pred} - I_{GT}  \right) \cdot m_{defect} \right]
 		}^2
	 \\  
	MaskedPSNR = -10 \cdot \log_{10}{\left(\frac{MSE_{mask}}{{MAX}^2} + {0.1}^8\right)}
\end{matrix}\right.,
\end{equation}
where $m_{defect}$ is defect binary mask(1:defects pixel, 0: flawless pixel), $I_{pred}$ and $I_{GT}$ are predicted restored and ground-truth frames.
We further use LPIPS\cite{zhang2018unreasonable}, BRISQUE\cite{mittal2012no}, NIQE\cite{mittal2012making} to capture perceptual quality.

\subsection{Restoration Results}
We report quantitative(\cref{tab:compare-sota-method}) and qualitative results for synthetic data and provide qualitative comparisons for real films due to copyright issue.

\begin{table}[t]
	\small
	\caption{\noindent \textbf{Quantitative comparisons with existing restoration methods on the synthetic dataset.}
	}
	\vspace{-0.1in}
	\label{tab:compare-sota-method}
	\centering
	\resizebox{1\linewidth}{!}{
		\begin{tabular}{c | c | c | c | c | c | c}
			\toprule 
			\multirow{2}{*}{Methods} & \multicolumn{6}{c}{
				mixed degradation / scratch degradation / noisy fast-motion film
			}  \\
			\cline{2-7} 
			& \textbf{MaskedPSNR}↑ & PSNR↑ & SSIM↑ & LPIPS↓ & BRISQUE↓ & NIQE↓\\
			\midrule
			
			Input &
			12.795 / 15.729 / 20.930 &
			22.552 / 26.771 / 36.929 &
			0.8135 / 0.9057 / 0.9852 &
			0.3694 / 0.2163 / 0.0471 &
			35.401 / 32.873 / 65.087 &
			10.505 / 9.421 / 8.857 \\
			
			DeepRemaster~\cite{iizuka2019deepremaster} &
			18.154 / 17.551 / 24.585 &
			22.962 / 25.483 / {\color{blue}31.026} &
			0.8322 / 0.8939 / 0.9716 &
			0.3318 / 0.2125 / 0.0745 &
			31.142 / 32.547 / 48.493 &
			10.183 / 9.204 / 7.733 \\
			
			RTN~\cite{wan2022bringing} &
			19.527 / 15.794 / \makebox[0pt][l]{\textemdash}\phantom{00.000} &
			22.700 / 21.900 / \makebox[0pt][l]{\textemdash}\phantom{00.000} &
			0.8295 / 0.8451 / \makebox[0pt][l]{\textemdash}\phantom{00.000} &
			0.1993 / 0.2052 / \makebox[0pt][l]{\textemdash}\phantom{00.000} &
			22.088 / 18.969 / \makebox[0pt][l]{\textemdash}\phantom{00.000} &
			10.574 / 10.119 / \makebox[0pt][l]{\textemdash}\phantom{0.000} \\
			
			RRTN~\cite{lin2024restoring} &
			21.644 / 17.134 / \makebox[0pt][l]{\textemdash}\phantom{00.000} &
			24.001 / 20.310 / \makebox[0pt][l]{\textemdash}\phantom{00.000} &
			0.8447 / 0.7935 / \makebox[0pt][l]{\textemdash}\phantom{00.000} &
			0.1729 / 0.1822 / \makebox[0pt][l]{\textemdash}\phantom{00.000} &
			15.437 / 15.317 / \makebox[0pt][l]{\textemdash}\phantom{00.000} &
			9.721 / 9.645 / \makebox[0pt][l]{\textemdash}\phantom{0.000} \\

			DeepEnhancer~\cite{jiang2024deepenhancer} &
			22.324 / 19.879 / 25.245 &
			24.881 / 21.888 / 22.922 &
			0.8618 / 0.8125 / 0.8946 &
			0.1537 / 0.1740 / 0.0927 &
			32.200 / 30.782 / 32.282 &
			10.209 / 10.066 / 6.743 \\
			
			\midrule
			\makecell[c]{RTN(finetuned \\
				with our data)~\cite{wan2022bringing}} &
			22.025 / 16.921 / 22.361 &
			24.823 / 24.078 / 27.520 &
			0.8773 / 0.8999 / 0.9377 &
			0.1441 / 0.1527 / 0.0816 &
			28.055 / 29.026 / 53.816 &
			10.252 / 9.575 / 7.631 \\
			
			\makecell[c]{RRTN(finetuned \\
				with our data)~\cite{lin2024restoring}} &
			{\color{blue}22.931} / 15.881 / 25.221 &
			{\color{blue}25.521} / 25.505 / 28.855 &
			0.8898 / 0.8993 / 0.9425 &
			0.1139 / 0.2116 / 0.0853 &
			21.276 / 29.904 / 49.564 &
			10.013 / 9.542 / 7.583 \\
			
			\midrule
			
			\makecell[c]{DIAMANT\\(commercial)~\cite{DIAMANT}} &
			19.661 / {\color{blue}26.771} / {\color{red}30.037} &
			{\color{red}25.774} / {\color{red}29.677} / {\color{red}33.092} &
			0.8637 / 0.9105 / 0.9824 &
			0.2682 / 0.1772 / 0.0419 &
			32.037 / 32.649 / 65.244 &
			9.150 / 9.486 / 8.917 \\

			\midrule
			
			Ours~ &
			{\color{red}24.580} / {\color{red}27.375} / {\color{blue}26.195} &
			25.464 / {\color{blue}25.806} / 25.042 &
			0.8414 / 0.8387 / 0.9541 &
			0.1307 / 0.1227 / 0.0963 &
			29.746 / 30.515 / 72.884 &
			9.641 / 9.786 / 8.905 \\
			\bottomrule
		\end{tabular}
	}
\end{table}

%
	%
	%
	%

\begin{figure}
	\centering
	\includegraphics[width=1.0\linewidth]{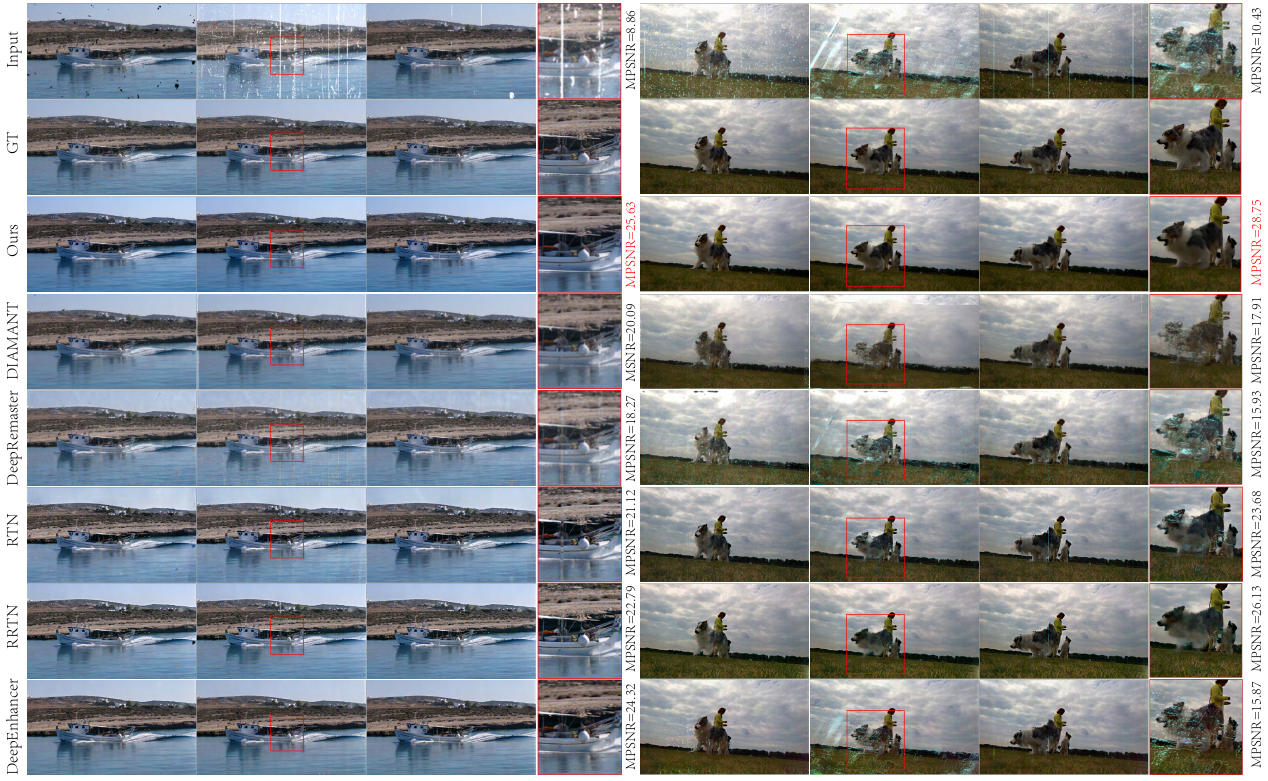}
	\caption{
		\textbf{Results on the mixed-degradation data}, depicting 3 consecutive frames(854*480 resolution), left:slow motion; right:fast motion.
	}
	\label{fig:synthetic_davis-mixed}
		\vspace{-0.5em}
\end{figure}

\noindent{\textbf{mixed degradation:}}
In Fig.\ref{fig:synthetic_davis-mixed},
DeepRemaster\cite{iizuka2019deepremaster}, RTN\cite{wan2022bringing}, RRTN\cite{lin2024restoring} have obvious defect remains and content-loss in restored results. DeepEnhancer\cite{jiang2024deepenhancer} shows slightly better defects restoration result, but would fail to restore fast-movement data. DIAMANT\cite{DIAMANT} maintains the best consistency of flawless region but shows distinguishable content-loss for fast-movement data.
In comparison, our approach show the best defect-restoring and structural consistency result on both fast and slow movement data.
Quantitatively, our approach outperforms existing methods on MaskedPSNR, observing 4.8dB gain over the commercial DIAMANT, which indicates our method better defect restoration ability. Though our method is constrained by VAE reconstruction loss and color-shifting issue, we observe the comparable PSNR and acceptable reconstruction loss. 

\begin{figure}
	\centering
	\includegraphics[width=1.0\linewidth]{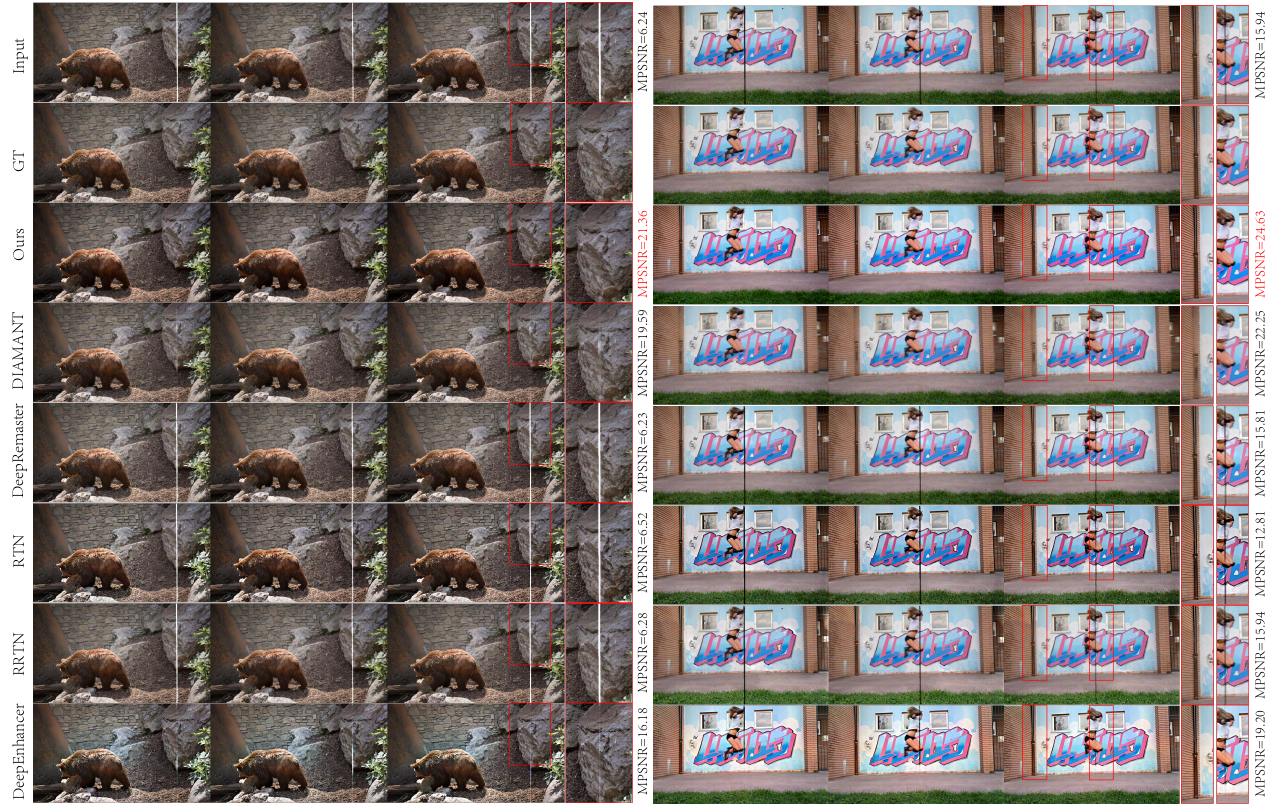}
	\caption{
		\textbf{Restoration results on the scratch degradation data}, depicting 3 consecutive frame(854*480 resolution),left:slow motion; right:fast motion.
	}
	\label{fig:synthetic_davis_scratch}
	\vspace{-0.5em}
\end{figure}

\noindent{\textbf{scratch degradation:}}
In Fig.\ref{fig:synthetic_davis_scratch}, DeepRemaster\cite{iizuka2019deepremaster}, RTN\cite{wan2022bringing}, RRTN\cite{lin2024restoring} completely fail for constant-scratch degradation. DeepEnhancer\cite{jiang2024deepenhancer} only restore partial scratch and introduce color flickering problem. DIAMANT\cite{DIAMANT} shows better result but introduce object disappearance.
Our approach shows the best scratch restoration result with strong structural consistency.
Quantitatively, our approach achieves the best MaskedPSNR and obtain 7dB gain on MaskedPSNR over open-source methods\cite{iizuka2019deepremaster,wan2022bringing, lin2024restoring, jiang2024deepenhancer}, which indicating strong scratch restoration ability.

\begin{figure}
	\centering
	\includegraphics[width=1.0\linewidth]{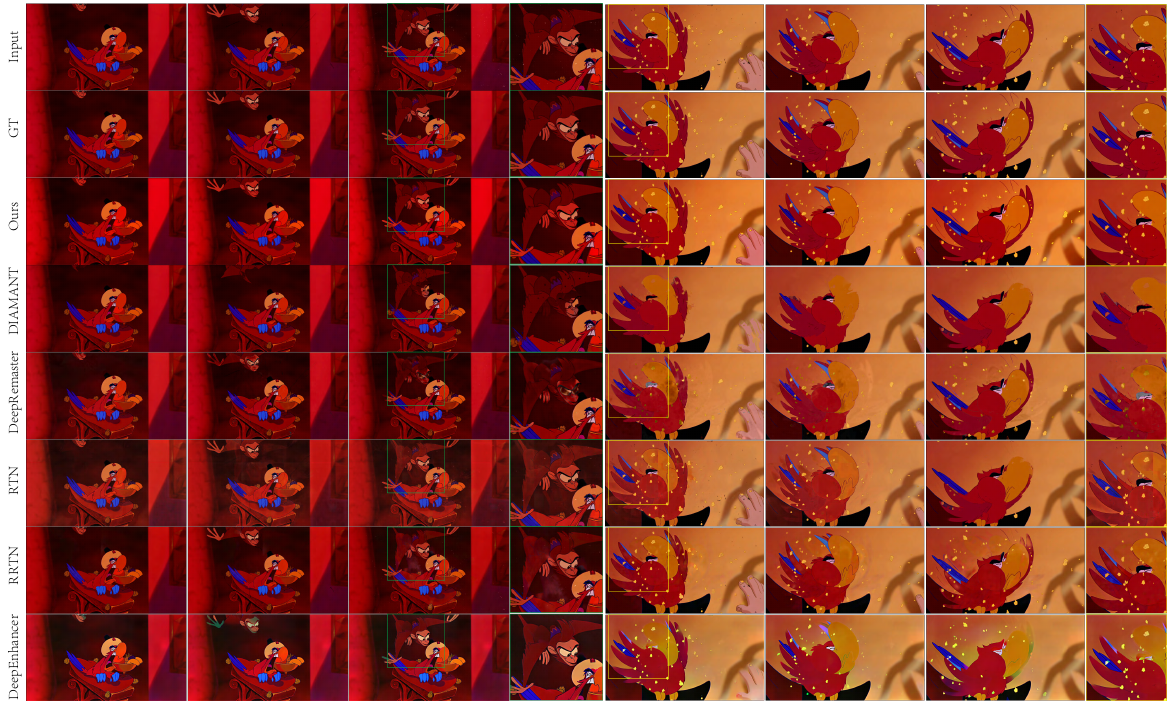}
	\caption{
		\textbf{Restoration results on the cartoon fast-motion data}, depicting 3 consecutive frame(1920*1040 resolution)
	}
	\label{fig:synthetic_cartoon}
		\vspace{-0.5em}
\end{figure}

\begin{figure}
	\centering
	\includegraphics[width=0.9\linewidth]{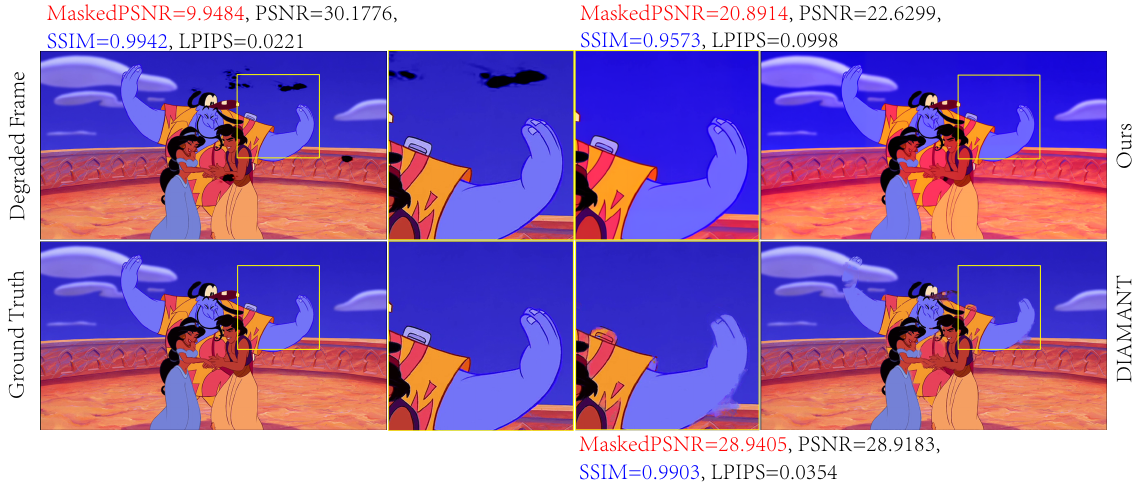}
	\caption{
		\textbf{Metrics Comparison of the fast-motion Cartoon:} Though slight color-shift issue results in sub-optimal metrics, our model maintains better structural consistency for fast-motion frames.
	}
	\label{fig:check_lowerPSNR_better_result}
	\vspace{-0.5em}
\end{figure}

\noindent{\textbf{noisy fast-motion film:}}
In \cref{fig:synthetic_cartoon}, our approach achieves the best defect-restoration result and content consistency, which make human post-processing possible. Though our method may suffer from the acceptable color-shifting issue as shown in Fig.\ref{fig:check_lowerPSNR_better_result}, which results in lower MaskedPSNR and PSNR metrics compared with DIAMANT\cite{DIAMANT}, our method maintains the best structural consistency without object disappearance.

\begin{figure}
	\centering
	\includegraphics[width=1\linewidth]{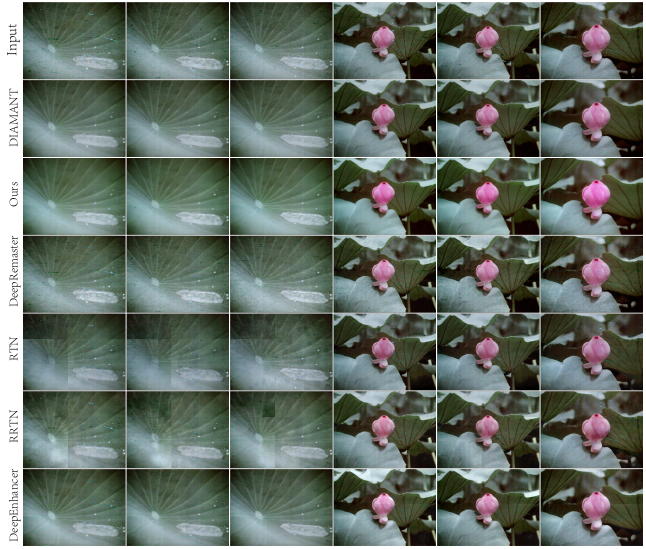}
	\caption{
		\textbf{Restoration results on the real films `Lotus'}, slow motion(2048*1536, 2K resolution)
	}
	\label{fig:real-lotus}
		\vspace{-0.5em}
\end{figure}

\noindent{\textbf{real film:}}
For slow-motion data(\cref{fig:real-lotus}), DIAMANT\cite{DIAMANT} prevail on defect removal and details reconstruction. Our HaineiFRDM delivers the second-best performance: it slightly blurs fine textures on the lotus leaf and introduces color shifts, but remove almost all the defects with minimal restoration artifacts, yielding the most expert-acceptable results compared with the open-source methods. 
For fast-motion data(\cref{fig:real-xiyangyang}), DIAMANT\cite{DIAMANT} completely fails and introduce body-distortion issue, which is unacceptable for post-processing. Our content-aware method successfully remove all the defects and maintain better overall restoration result.


\subsection{Ablation Study}

\begin{figure}
	\centering
	\includegraphics[width=1.0\linewidth]{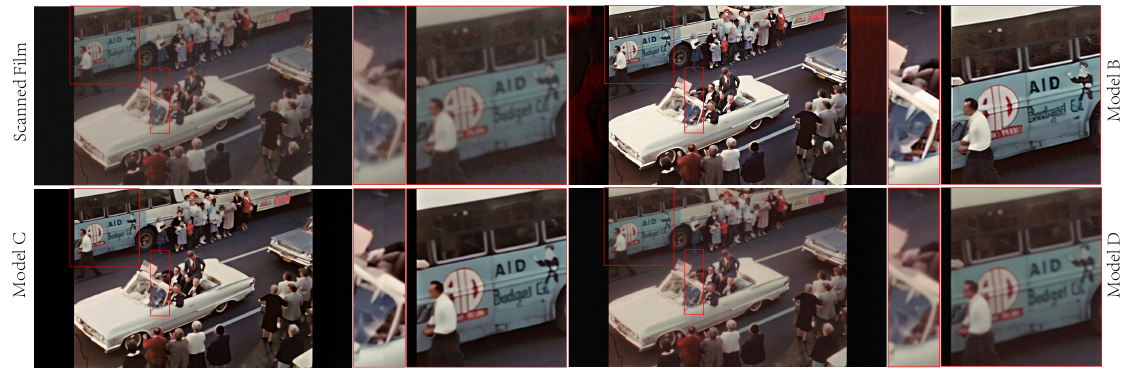}
	\caption{\textbf{Ablation study:} Qualitative Comparison of our proposed modules.}
	\label{fig:ablation-freqModule}
	\vspace{-0.5em}
\end{figure}

To verify the e effectiveness of our proposed modules, this section presents ablation studies on DAVIS-based synthetic degradation data and real-film. Observing \cref{tab:ablation-modules}, the model with global-frame module obtaining increase in all metrics, indicating value of global information. In \cref{fig:ablation-freqModule}, the global prompt module avoid the misleading local prompts and prevent the illusion effect in dark background. And frequency-based module helps increase the consistency of textures on the bus.
\begin{table}
	\caption{\textbf{Ablation study:} Quantitative Comparison of our proposed modules.}
	\label{tab:ablation-modules}
	
	\centering
	\small
	\setlength{\tabcolsep}{3pt} 
	\renewcommand{\arraystretch}{1.1}
	\resizebox{1\linewidth}{!}{
		\begin{tabular}{
				c
				@{\hspace{2pt}}c@{\hspace{2pt}}
				c@{\hspace{2pt}}
				c@{\hspace{2pt}}
				c@{\hspace{2pt}}
				c
				|cccc
			}
			\hline\hline
			\textbf{Models} &
			\makecell[c]{Local-\\Prompt\\Module} &
			\makecell[c]{Global-\\Visual\\Module} &
			\makecell[c]{Global-\\Prompt\\Module} &
			\makecell[c]{Texture\\Module} &
			\makecell[c]{Tuning\\Temporal-\\Module} &
			\textbf{MaskedPSNR} &
			\textbf{PSNR} &
			\textbf{SSIM} &
			\textbf{LPIPS} \\
			\hline\hline
			{Model A} & \ding{51} & - & - & - & - & 24.1348 & 24.2054 & 0.8008 & 0.1390 \\
			{Model B} & \ding{51} & \ding{51} & - & - & - & 25.1190 & 25.3210 & 0.7881 & 0.1387 \\
			{Model C} & - & \ding{51} & \ding{51} & - & - & 25.6743 & 25.3953 & 0.7991 & 0.1413 \\
			{Model D} & - & \ding{51} & \ding{51} & \ding{51} & - & 25.8112 & 25.7258 & \textbf{0.8221} & 0.1126 \\
			{HaineiFRDM} & - & \ding{51} & \ding{51} & \ding{51} & \ding{51} & \textbf{26.6190} & \textbf{26.3980} & 0.8201 & \textbf{0.1076} \\
			\bottomrule
		\end{tabular}
	}
\end{table}

 \begin{figure}
	\centering
	\includegraphics[width=1\linewidth]{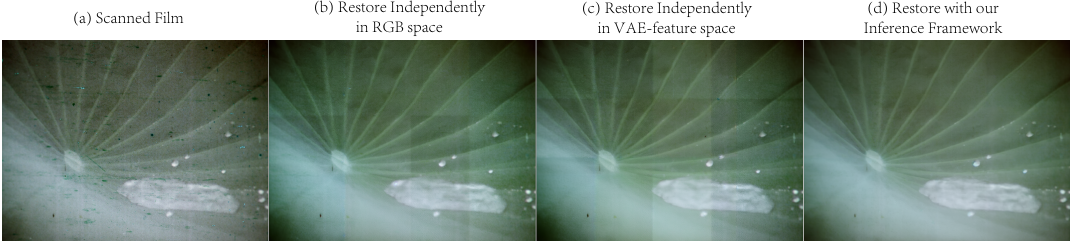}
	\caption{
		\textbf{Ablation study of our patch-consistent inference framework:} (a) 2K-resolution scanned film; (b) we split patches in RGB space and restore each patch independently; (c) we split patches in VAE feature space, restore each patch independently, fuse patches back to whole frame feature and decode the feature into RGB frame; (d) use our inference framework to restore every patches in cooperation.}
	\label{fig:exp-ablation-global-residua}
	\vspace{-0.5em}
\end{figure}

\section{Conclusion and Limitation}
\noindent{\textbf{Conclusion:}} We propose a diffusion-based framework for high-resolution film restoration. Extensive experiments demonstrate the strong structural consistency and superior restoration quality on both synthetic and real films.

\noindent{\textbf{Limitation:}} 1) Since the restoration is performed in the latent space, VAE-reconstruction loss may introduce slight color shifts and slight texture smoothing, as shown in \cref{fig:check_lowerPSNR_better_result};
2) Inference is relatively slow on an RTX 4090 GPU and may require processing multiple clips in parallel across GPUs for faster restoration;
3) The cross-scene contents have not been explored by our model yet.

%
%



%
%
\bibliographystyle{splncs04}
\bibliography{main}

\clearpage


\end{document}